\documentclass{article}

\usepackage[preprint]{neurips_2026}

\usepackage[utf8]{inputenc} 
\usepackage[T1]{fontenc}    
\usepackage{hyperref}       
\usepackage{url}            
\usepackage{booktabs}       
\usepackage{amsfonts}       
\usepackage{nicefrac}       
\usepackage{microtype}      
\usepackage{xcolor}         
\usepackage{graphicx}
\usepackage{subcaption}
\title{Cross-View Urban Traffic Dataset:\\
Drone-Supervised Ground Truth for Monocular Bird's-Eye View Localization}

\author{%
  Prakhar Bhardwaj \\
  Faculty of Computer Science and Mathematics, OTH Regensburg \\
  \texttt{prakhar.bhardwaj@oth-regensburg.de}
  \And
  Simone Weikl \\
  Faculty of Computer Science and Mathematics, OTH Regensburg \\
  \texttt{simone.weikl@oth-regensburg.de}
  \AND
  Kilian Mang \\
  Faculty of Computer Science and Mathematics, OTH Regensburg \\
  \texttt{kilian.mang@st.oth-regensburg.de}
  \And
  Elia Jonas Sandtner \\
  Faculty of Computer Science and Mathematics, OTH Regensburg \\
  \texttt{elia.sandtner@st.oth-regensburg.de}
}

\usepackage{amsmath}

\begin{document}

\maketitle

\begin{abstract}
We introduce a dataset and benchmark for cross-view urban traffic perception built from synchronized ego-centric bicycle videos and aerial drone videos recorded at real urban intersections. The benchmark targets two linked tasks: cross-view identity matching between street-view and drone-view object tracks, and ego-to-bird's-eye-view prediction using aerial supervision. In contrast to prior urban driving and V2X datasets, our benchmark provides identity-level alignment across radically different viewpoints together with standardized evaluation, annotation tooling, and baseline implementations. This setting is motivated by intersection-centric traffic analysis, where identity preservation, local interactions, and global spatial structure must be reasoned about jointly across views. We evaluate methods at both the track and frame levels, including cross-view ID precision/recall/IDF1, near--far breakdowns, temporal stability, and consistency metrics. We also provide baseline results for wedge-based cross-view matching and for three BEV prediction baselines: inverse perspective mapping, a MonoLayout-style learned baseline, and a regression baseline. The results show that the benchmark is feasible but challenging: cross-view matching achieves strong recall yet remains limited by over-assignment and temporal inconsistency, while ego-to-BEV prediction benefits from aerial supervision but remains far from saturated under lightweight monocular sensing. We hope that this benchmark will support future research on cross-view perception, urban scene alignment, and ego-to-global traffic understanding.
\end{abstract}

\section{Introduction}
Progress in urban perception depends not only on better models, but also on better evaluation resources. Benchmarks determine which capabilities are measured, which failure modes are exposed, and which research questions become scientifically actionable \cite{coco,imagenet}. In traffic perception, recent datasets have advanced cooperative perception, bird's-eye-view (BEV) reasoning, and multi-view scene understanding, especially in vehicle-to-infrastructure and multi-agent settings. Representative examples include V2X-Seq~\cite{yu2023v2xseqlargescalesequentialdataset}, TUMTraf-V2X~\cite{zimmer2024tumtraf}, and UrbanIng-V2X~\cite{urbaningv2x2025}.

However, an important evaluation gap remains: existing urban datasets rarely provide identity-level alignment across radically different viewpoints, particularly between an ego-centric street view and an aerial drone view over the same intersection. Most cooperative perception datasets emphasize sensor fusion, 3D detection, or forecasting, rather than persistent cross-view identity correspondence between local ego observations and global aerial context~\cite{yu2023v2xseqlargescalesequentialdataset,zimmer2024tumtraf,urbaningv2x2025}. Related cross-view benchmarks in geo-localization, such as VIGOR~\cite{vigor}, operate primarily at the image level and do not provide synchronized video, object-track correspondences, or metric traffic-participant positions. This missing link matters because the ego view and the aerial view provide complementary information. The street view captures appearance, motion cues, and local interaction context, but is highly affected by blur, occlusion, ego motion, and track fragmentation. The aerial view provides global spatial structure and cleaner scene geometry, but weaker appearance detail and a fundamentally different viewpoint. A system that performs well in only one of these views may still fail to preserve identity across them, maintain temporal consistency, or recover a coherent traffic layout.

Intersections are a particularly important setting for this problem. They concentrate the traffic events that matter most for urban analysis: merging, crossing, yielding, turning conflicts, short-term interactions between vehicles and vulnerable road users, and local density changes. As a result, they are a natural test bed for use cases such as intersection monitoring, traffic interaction analysis, local density estimation, queue and occupancy analysis, and understanding of the ego-to-global scene. These use cases require not only detecting objects, but also linking identities across viewpoints and reasoning about their spatial arrangement in a shared frame. At the same time, monocular BEV methods have shown that recovering scene structure from ego-centric imagery is feasible. Methods such as MonoLayout~\cite{monolayout} demonstrated the promise of single-image BEV reasoning, while broader BEV perception literature has established BEV as a useful common representation for downstream urban reasoning~\cite{ma2023visioncentricbevperceptionsurvey,yazgan2024collaborativeperceptiondatasetsautonomous}. Yet current BEV benchmarks are typically not paired with explicit cross-view identity supervision from aerial observations.

\begin{figure}[!h]
    \centering
    \includegraphics[width=\textwidth]{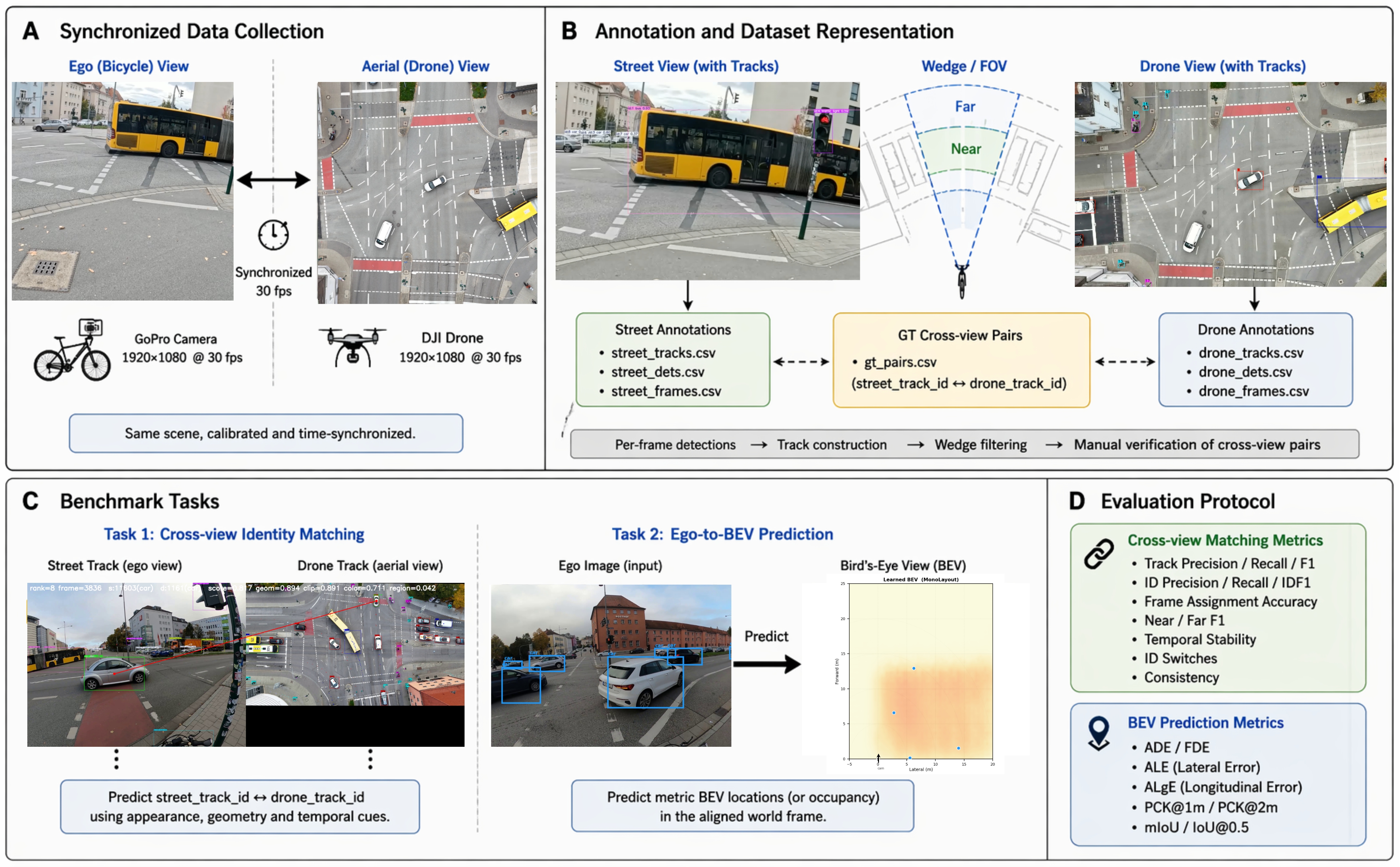}
    \caption{Overview of the proposed cross-view urban traffic dataset and benchmark. (\textbf{A}) Each scene contains synchronized ego-centric bicycle video and aerial drone video over the same urban intersection. (\textbf{B}) From these views, we derive tracked annotations, wedge-filtered representations, and manually verified cross-view correspondences. (\textbf{C}) The benchmark supports two tasks: cross-view identity matching and ego-to-BEV prediction map with aerial supervision. (\textbf{D}) Evaluation uses track-level, frame-level, geometric, and temporal metrics, and BEV localization metrics.}
    \label{fig:dataset_overview}
\end{figure}

In this paper, we introduce a cross-view urban traffic dataset and benchmark built from synchronized ego-centric bicycle videos and aerial drone videos recorded at real urban intersections (Figure~\ref{fig:dataset_overview}). The benchmark targets two linked tasks: \emph{cross-view identity matching}, where the goal is to associate street-view object tracks with their corresponding aerial tracks, and \emph{ego-to-BEV prediction}, where the goal is to recover BEV traffic structure from ego-centric observations using aerial supervision. Unlike prior cooperative perception datasets, our formulation explicitly connects local identity-rich ego observations with global aerial spatial context at the level of traffic participants and trajectories~\cite{urbaningv2x2025,zimmer2024tumtraf,yu2023v2xseqlargescalesequentialdataset}. Beyond the data itself, we provide the evaluation methodology needed to make the resource scientifically useful. For cross-view matching, we evaluate both track-level and frame-level performance, including cross-view ID precision/recall, IDF1, frame assignment accuracy, near--far breakdowns, temporal stability, and consistency. For ego-to-BEV prediction, we provide aligned supervision and baseline protocols for both geometric and learning-based methods. To support reproducibility, we release the benchmark pipeline, including wedge-based region extraction, crop generation, embedding computation, a frame-batch annotation interface, baseline implementations, and evaluation scripts. The dataset, annotation tools, preprocessing pipeline, and evaluation code will be made publicly available at \url{https://huggingface.co/datasets/prakharbh/CrossViewUrbanTrafficDataset} and the codebase is available at \url{https://github.com/oth-aifiud/Cross-View-Urban-Traffic-Dataset}.

Our experiments show that the benchmark exposes meaningful sources of difficulty, including viewpoint shift, street-view track fragmentation, occlusion, temporal inconsistency, and near--far asymmetry, while remaining tractable enough to support measurable progress. We therefore view this resource as a foundation for future research on cross-view perception, urban scene alignment, and ego-to-global traffic reasoning.

We make the following contributions:
\begin{enumerate}
  \item We introduce cross-view urban traffic dataset \textbf{CVUTD} with identity-aligned supervision across synchronized ego and aerial views, recorded at 8 urban intersections in Regensburg, Germany.
  \item We define a benchmark with two linked tasks: cross-view identity matching and ego-to-BEV prediction.
  \item We formalize an evaluation protocol spanning track-level, frame-level, geometric, and temporal metrics.
  \item We release a reproducible toolkit for preprocessing, annotation, baseline evaluation, and future benchmarking.
\end{enumerate}

\section{Related Work}

\paragraph{Urban driving and monocular BEV datasets.}
Large-scale autonomous driving datasets such as \emph{KITTI}, \emph{Argoverse}, and \emph{nuScenes} have driven progress in 3D perception and BEV reasoning, but they rely on sensor-rich vehicles equipped with LiDAR, GPS/IMU, or multi-camera rigs~\cite{kitti,argoverse,nuscenes}. KITTI and nuScenes provide strong geometric supervision through LiDAR-based annotations, while Argoverse further adds map priors and tracking/forecasting benchmarks~\cite{kitti,argoverse,nuscenes}. More recent BEV-oriented methods such as \emph{BEVerse} similarly assume calibrated multi-camera setups~\cite{beverse}. These assumptions do not hold in our target setting: a lightweight bike-mounted monocular camera operating in real urban intersections under strong compute and sensing constraints. Our dataset targets this underexplored regime and provides metric BEV supervision using synchronized aerial observations.

\paragraph{Cross-view datasets and geo-localization.}
A related line of work studies cross-view matching between street-level and aerial imagery, primarily for geo-localization. Datasets such as \emph{CVUSA} and \emph{VIGOR} pair ground and aerial images at scale and demonstrate the difficulty of cross-view retrieval under substantial viewpoint change~\cite{cvusa,vigor}. However, these benchmarks are formulated at the \emph{image level}, typically using static aerial or satellite imagery, and do not provide synchronized video, object-level identity correspondences, or metric traffic-participant positions~\cite{cvusa,vigor}. Other cross-view datasets such as \emph{DroneVehicle} pair drone and ground imagery for detection, but not synchronized street-drone video with metric BEV supervision for urban traffic participants~\cite{dronevehicle}. In contrast, our dataset targets \emph{video-synchronized, metric-calibrated, object-level} cross-view correspondence, with explicit support for both identity matching and ego-to-BEV prediction.

\paragraph{IPM and learned BEV projection.}
Classical monocular BEV projection is commonly based on inverse perspective mapping (IPM), which assumes a ground plane and calibrated camera geometry. While simple and efficient, IPM is brittle under non-planarity, calibration error, and occlusion. Learning-based alternatives such as \emph{MonoLayout} infer BEV structure directly from monocular observations, while more recent methods such as \emph{Lift-Splat-Shoot}, \emph{BEVDet}, and \emph{BEVFormer} learn stronger BEV representations using depth reasoning, multi-camera fusion, or spatiotemporal transformers~\cite{monolayout,lss,bevdet,bevformer}. However, these methods are generally developed and benchmarked on datasets with richer supervision and sensor setups than are available in our bike-camera setting~\cite{monolayout,lss,bevdet,bevformer}. Our benchmark complements this literature by providing a lower-cost path to metric BEV supervision through synchronized drone observations.

\paragraph{Drone-assisted perception and supervision.}
Drones have also been widely used as perception platforms in urban and aerial vision datasets. For example, \emph{VisDrone} and UAV-based vehicle-detection benchmarks study detection and tracking from above~\cite{visdrone,uavvehicle}. In these works, however, the drone is itself the sensing platform of interest rather than a \emph{supervision source} for street-level perception. Our use of the aerial view is different: the drone provides global spatial context and metric supervision for an ego-centric street-view benchmark, enabling the same synchronized scene to support both cross-view identity matching and BEV prediction.

\paragraph{Positioning of our benchmark.}
Taken together, prior work provides strong resources for cooperative perception, cross-view retrieval, monocular BEV reasoning, and aerial traffic perception, but these capabilities remain largely disjoint. To the best of our knowledge, there is still no benchmark that jointly supports \emph{(i)} cross-view identity matching between synchronized ego and aerial traffic observations and \emph{(ii)} ego-to-BEV prediction under aerial metric supervision in real urban intersections.

\section{Dataset}

We introduce a cross-view urban traffic dataset built from synchronized \emph{ego-centric bicycle videos} and \emph{aerial drone videos} recorded at real urban intersections. Unlike prior datasets that focus on infrastructure-assisted cooperative perception or BEV prediction in sensor-rich vehicles, our dataset targets a lightweight bike-camera setting in which global traffic structure must be recovered from a monocular ego view and aligned with an aerial reference.

\subsection{Recording Setup}

The ego-centric view is recorded using a GoPro Hero 11 mounted on the bicycle handlebar at approximately $1.1$m above the ground. Videos are captured at 4K resolution ($3840 \times 2160$) and 30 fps in linear mode. Camera intrinsics are estimated using OpenCV checkerboard calibration, yielding $f_x = 2060.4$, $f_y = 2057.4$, $c_x = 1905.9$, and $c_y = 1079.4$ in pixel units. The camera pitch angle is approximately $-8^\circ$ below the horizontal. The aerial view is recorded using a consumer quadrotor drone hovering at approximately $60$m above the intersection center. Drone videos are also captured at 4K resolution and 30 fps. At this altitude, the drone covers roughly $108 \times 61$m of ground area under an approximately $84^\circ$ horizontal field of view, which is sufficient to capture the full spatial extent of a typical urban intersection together with its approach lanes. Street-view and drone-view videos are synchronized manually by aligning a common visible event, such as a traffic light transition or a hand gesture, to within approximately $\pm 1$ frame (33 ms at 30 fps).

\subsection{Data Collection and Annotation}

We record data at eight urban intersections in Regensburg, Germany, spanning residential, commercial, and arterial road types. Each scene consists of one continuous bicycle traversal of an intersection while the drone observes the full traffic layout from above. This setup captures synchronized local and global views of the same traffic event, including vehicles, cyclists, pedestrians, and public transport. The dataset is organized around \emph{track-level object annotations} in both views. Street-view annotations contain one row per detection per frame with frame index, track ID, class, and bounding box information. Drone-view annotations additionally provide metric world coordinates, which define the reference frame used throughout the benchmark. From these per-view annotations, we construct human-verified cross-view correspondences in the form of \texttt{gt\_pairs.csv}. To make annotation scalable, we use a web-based interface that supports both single-track and frame-batch verification. Additional details on the annotation process and anonymization are provided in Appendix~\ref{app:annotation}.

\subsection{Benchmark Tasks}

The dataset supports two benchmark tasks.
\paragraph{Task A: Cross-view identity matching.}
Given synchronized street-view and drone-view object tracks, the goal is to recover the correspondence between a street-view track and the matching drone-view track. The task is evaluated at both the frame level and the track level, and is challenging because of viewpoint differences, occlusion, temporal fragmentation, and ambiguity among visually similar traffic participants.

\paragraph{Task B: Ego-to-BEV prediction.}
Given an ego-centric street-view image or sequence, the goal is to predict the spatial arrangement of traffic participants in bird's-eye-view coordinates. The drone view provides aligned metric supervision for this task through world-coordinate annotations and cross-view correspondences. This setting supports both geometric and learned BEV baselines.

By coupling these two tasks, the dataset enables evaluation of both \emph{who is where} across viewpoints and \emph{where everyone is} in a shared BEV frame.

\subsection{Dataset Statistics}

Table~\ref{tab:dataset_stats} summarizes the current release. The dataset contains 8 urban intersections, 432{,}000 street-view frames, 144{,}000 drone-view frames, 3{,}847 street-view tracks, 2{,}103 drone-view tracks, and 1{,}891 validated cross-view correspondences. Because street-view tracking is substantially more fragmented than drone-view tracking, the number of unique street-view track IDs is larger; this should be interpreted as a property of ego-centric tracking difficulty rather than a difference in the number of real traffic participants.

\begin{table}[t]
  \caption{Summary statistics of the proposed cross-view urban traffic dataset. The dataset spans 8 urban intersections and contains nearly 1.9K matched cross-view track correspondences for identity matching and BEV prediction under synchronized ego and aerial views.}
  \label{tab:dataset_stats}
  \centering
  \begin{tabular}{lr}
    \toprule
    Statistic & Value \\
    \midrule
    Scenes (intersections)    & 8 \\
    Street frames (total)     & 432{,}000 \\
    Drone frames (total)      & 144{,}000 \\
    Street tracks             & 3{,}847 \\
    Drone tracks              & 2{,}103 \\
    Matched cross-view tracks & 1{,}891 \\
    Overall match rate        & 49.2\% \\
    Street detections         & 187{,}432 \\
    Drone detections          & 94{,}218 \\
    \bottomrule
  \end{tabular}
\end{table}
The benchmark combines substantial scale with nontrivial cross-view supervision while preserving realistic challenges such as ego-view track fragmentation, occlusion, and scene-to-scene variability. Additional per-class statistics are reported in Appendix~\ref{app:dataset_stats}.

\section{Evaluation Protocol}

A core contribution of the benchmark is a standardized evaluation protocol for two complementary tasks: \emph{cross-view identity matching} and \emph{ego-to-BEV prediction}. For cross-view matching, we evaluate both track-level correctness and frame-level temporal consistency, together with distance-sensitive and temporal diagnostics that expose key sources of difficulty such as viewpoint ambiguity, near--far asymmetry, and identity instability. For ego-to-BEV prediction, we evaluate spatial accuracy in a shared metric frame defined by the drone annotations. This emphasis on explicit evaluation is aligned with benchmark-oriented practice in tracking, trajectory prediction, and BEV reasoning~\cite{motchallenge,ristani2018mtmct,monolayout}.

\subsection{Cross-view Identity Matching}

Cross-view identity matching is evaluated at both the \emph{track level} and the \emph{frame level}. At the track level, a method predicts a mapping from each street-view track to a drone-view track, and we report precision, recall, and F1 against the human-verified ground-truth correspondences. At the frame level, we evaluate whether the assigned drone identity is correct in each temporally aligned frame in which a ground-truth street track is visible. We report cross-view ID precision (IDP), recall (IDR), IDF1, and frame assignment accuracy. These frame-level metrics are especially important in our setting because a method may recover the correct correspondence at the track level while still exhibiting poor temporal behavior, for example by switching identities or matching correctly only in a subset of frames.

\subsection{Near--Far and Temporal Diagnostics}

To expose the geometric dependence of cross-view matching, we divide GT-visible frames into \emph{near} and \emph{far} subsets based on drone-side metric distance, with a threshold of 30\,m in our experiments. We then report frame-level matching performance separately in the two regimes. This breakdown makes explicit a central property of the task: matching difficulty depends strongly on the geometric regime in which an object is observed. Because the benchmark is track-based, temporal behavior is also a first-class aspect of evaluation. In addition to frame-level IDF1, we report three temporal diagnostics: \emph{stability}, which measures how often the correct drone identity is assigned over the lifespan of a street-view track; \emph{ID switches}, which count changes in the predicted drone identity over time; and \emph{consistency}, which measures how dominant the most frequent predicted identity is within a track. Together, these metrics reveal whether a method is merely correct occasionally or remains coherent over time. Full metric definitions are provided in Appendix~\ref{app:matching_results}.

\subsection{Ego-to-BEV Prediction}

The second task evaluates whether an ego-centric street-view model can recover the spatial arrangement of traffic participants in a common bird's-eye-view frame. The drone annotations provide the metric world coordinates used for supervision and evaluation. Depending on the baseline, a method may output either explicit object coordinates or a rasterized BEV occupancy representation. We therefore support both \emph{coordinate-level} and \emph{occupancy-level} metrics. For coordinate prediction, we report average displacement error (ADE), final displacement error (FDE), lateral error (ALE), longitudinal error (ALgE), and thresholded localization accuracy such as PCK@1m and PCK@2m~\cite{adefde,criteria,pck,yangramanan}. For occupancy prediction, we report mIoU together with thresholded overlap measures such as IoU@0.25 and IoU@0.5~\cite{monolayout,iousegsurvey}. This formulation allows geometric baselines and learned BEV baselines to be evaluated under a common protocol.

\subsection{Splits and Benchmark Use}

We recommend scene-level train/validation/test splits rather than frame-level splits in order to avoid temporal leakage and over-optimistic generalization estimates. All evaluation is performed scene-wise, and global metrics are obtained by aggregating over the relevant scenes. Because the benchmark is intended to study generalization across intersections, scene-level separation is essential. The benchmark therefore supports two complementary modes of use: \emph{diagnostic evaluation}, where methods are analyzed scene by scene and broken down by class, distance regime, and temporal behavior; and \emph{leaderboard evaluation}, where global aggregate metrics are reported across predefined test scenes.

\section{Baselines}

To make the benchmark immediately usable, we provide baseline methods for both benchmark tasks: \emph{cross-view identity matching} and \emph{ego-to-BEV prediction}. These baselines are intended to calibrate task difficulty and to establish reproducible reference points for future work. Additional implementation details are provided in Appendix~\ref{app:implementation}.

\subsection{Cross-view Identity Matching}

For cross-view identity matching, we implement a wedge-based baseline that combines appearance similarity, geometric compatibility, and temporal voting. The method operates on synchronized street-view and drone-view detections restricted to an ego-centric wedge region, reducing irrelevant clutter while preserving the traffic participants most relevant to cross-view matching. Street-view crops are compared against drone-view candidates from the synchronized frame using CLIP-based appearance embeddings~\cite{clip}, augmented with angular consistency and rank consistency cues. Matching is performed using a two-pass Hungarian assignment, with a more appearance-driven near regime and a more geometry-driven far regime, and the resulting frame-level assignments are stitched into track-level correspondences by confidence-aware temporal voting.

\subsection{Ego-to-BEV Prediction}

For ego-to-BEV prediction, we evaluate three baseline families spanning geometric and learned approaches. The first is a geometric inverse perspective mapping (IPM) baseline, which projects monocular image observations onto a ground plane using camera calibration and a planar-world assumption, and serves as a simple lower-bound reference~\cite{ipm}. The second is a MonoLayout-style learned BEV baseline~\cite{monolayout}, which predicts a rasterized BEV occupancy representation from the ego-view image using aerially aligned supervision. The third is a lightweight regression-based baseline that predicts metric BEV coordinates directly from street-view object observations, providing an object-centric alternative to dense occupancy prediction. Together, these baselines span three complementary assumptions: pure geometry, dense learned BEV reasoning, and lightweight object-level localization. This allows the benchmark to expose which aspects of performance are limited primarily by viewpoint gap, geometric assumptions, or sparse supervision.

\section{Experiments}

We evaluate the proposed benchmark on the two tasks introduced in Section~3 under the protocol defined in Section~4. These experiments are intended not only to compare baseline methods, but also to demonstrate how the dataset supports cross-view identity analysis and ego-to-global traffic reasoning at urban intersections.

\subsection{Experimental Setup}

For cross-view identity matching, we evaluate the wedge-based baseline described in Section~5 using the manually verified \texttt{gt\_pairs.csv} annotations. We report track-level precision/recall/F1, frame-level ID precision/recall/F1, frame assignment accuracy, near--far breakdowns, stability, ID switches, and consistency. For ego-to-BEV prediction, we evaluate three baselines: a geometric inverse perspective mapping (IPM) baseline~\cite{ipm}, a MonoLayout-style learned BEV predictor~\cite{monolayout}, and a regression-based baseline. Coordinate-level baselines are evaluated using ADE, FDE, ALE, ALgE, and PCK@1m / PCK@2m, while occupancy-style baselines are evaluated using mIoU and thresholded IoU metrics. We report BEV results on scenes with valid coordinate alignment. All experiments use scene-level splits to avoid temporal leakage between train, validation, and test scenes.

\subsection{Cross-view Matching Results}

Table~\ref{tab:matching_results} reports the main matching results. The wedge-based baseline achieves perfect track-level recall, indicating that it can recover nearly all ground-truth correspondences at least once, but track-level precision remains substantially lower (36.0\%), yielding a track-level F1 of 52.9\%. This shows that over-assignment and ambiguous correspondences remain major failure modes. At the frame level, the baseline achieves 75.2\% ID precision, 34.0\% ID recall, and 46.8\% IDF1, with a frame assignment accuracy of 30.6\%. The gap between track-level recall and frame-level recall highlights a central challenge of the benchmark: recovering a correspondence at least once is much easier than maintaining it correctly over time. Qualitative matching results are provided in Appendix~\ref{app:matching_results}

\begin{table}[!h]
  \caption{Cross-view identity matching results on the proposed benchmark. The baseline achieves strong recall but leaves substantial room for improvement in precision and temporal completeness.}
  \label{tab:matching_results}
  \centering
  \small
  \begin{tabular}{lccccccc}
    \toprule
    Method & Track P & Track R & Track F1 & IDP & IDR & IDF1 & FrameAcc \\
    \midrule
    Wedge matching baseline & 36.0 & 100.0 & 52.9 & 75.2 & 34.0 & 46.8 & 30.6 \\
    \bottomrule
  \end{tabular}
\end{table}

\subsection{Near--Far and Temporal Analysis}

Table~\ref{tab:matching_breakdown} reports the distance-sensitive and temporal diagnostics. Interestingly, performance is stronger in the far regime (66.1 F1) than in the near regime (41.3 F1). Although nearby objects are visually larger, they also appear in denser and more cluttered traffic configurations, where multiple similar objects compete for the same aerial neighborhood. Temporal metrics reveal that the baseline remains only moderately stable over the lifespan of a track. Mean stability is 34.0, while the mean number of ID switches is 0.491 per track. Consistency remains high at 95.3, suggesting that once the method converges to a dominant correspondence, that correspondence is usually coherent. Together, these results show that the benchmark difficulty lies not only in finding a correct match, but in maintaining it robustly over time.

\begin{table}[!h]
  \caption{Diagnostic breakdown of cross-view matching performance. The table reports distance-sensitive and temporal metrics, showing how difficulty varies across geometric regimes and over the lifespan of a track.}
  \label{tab:matching_breakdown}
  \centering
  \small
  \begin{tabular}{lccccc}
    \toprule
    Method & Near F1 & Far F1 & Stability & ID switches & Consistency \\
    \midrule
    Wedge matching baseline & 41.3 & 66.1 & 34.0 & 0.491 & 95.3 \\
    \bottomrule
  \end{tabular}
\end{table}

\subsection{Ego-to-BEV Results}

Tables~\ref{tab:bev_coord_results} and~\ref{tab:bev_occ_results} report the BEV baseline results on the aligned evaluation split. The geometric IPM baseline provides a weak but interpretable reference point, with an ADE of 5.43\,m and only 6.7\% PCK@2m, confirming that direct ground-plane projection from monocular detections is insufficiently robust in the bike-mounted setting.

\begin{figure*}[!h]
  \centering
  \includegraphics[width=\textwidth]{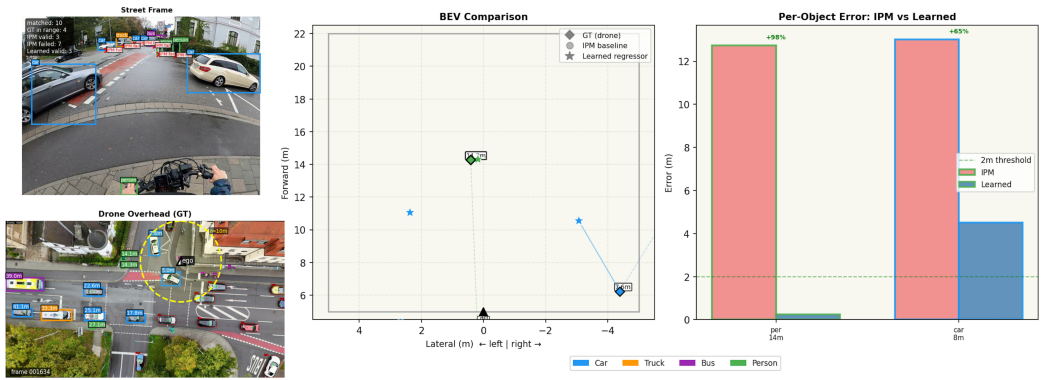}
  \caption{Qualitative example of the BBox BEV regressor baseline. From left to right: the ego street-view frame, BEV comparison of drone ground truth, IPM, and learned regression outputs, per-object error comparison, and the synchronized drone view. The learned regressor improves several predictions over raw IPM, but substantial errors remain in difficult cases.}
  \label{fig:bbox_regressor_vis}
\end{figure*}

The BBox BEV regressor improves over IPM, reducing ADE from 5.43\,m to 4.45\,m, corresponding to an 18.0\% improvement. It achieves an average lateral error of 1.68\,m and longitudinal error of 3.77\,m, with PCK@1m and PCK@2m of 3.5\% and 13.0\%, respectively. These results show that lightweight learning on top of explicit geometric cues is beneficial, though still limited by class imbalance and viewpoint variation. Figure~\ref{fig:bbox_regressor_vis} provides a representative qualitative example: compared with raw IPM, the learned regressor often shifts predictions closer to the drone-supervised BEV positions. Some raw IPM projections lie outside the displayed BEV window, reflecting the large localization errors that occur in difficult cases.

\begin{figure*}[!h]
  \centering
\includegraphics[width=\linewidth,keepaspectratio]{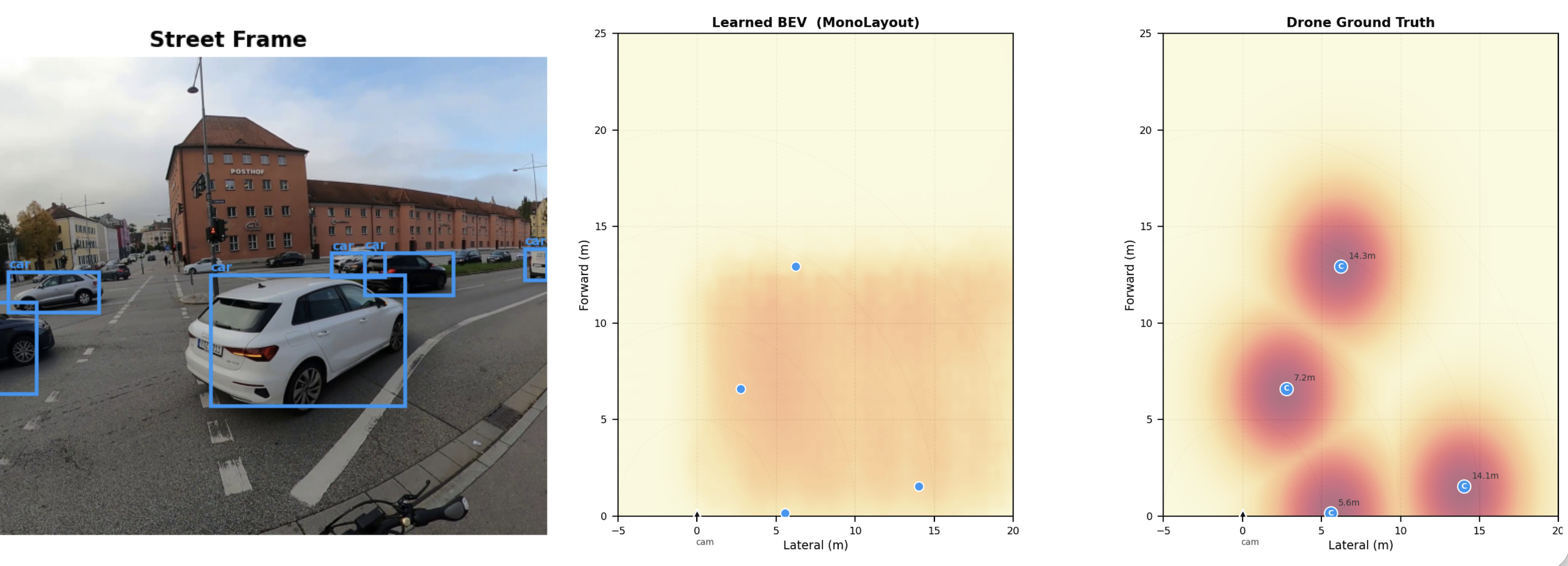}
  \caption{Qualitative visualization of the MonoLayout-style learned BEV baseline. From left to right: the ego street-view frame, the learned BEV map, and the drone-derived BEV ground truth. In the BEV map, warmer colors indicate higher predicted occupancy confidence, while cooler colors indicate lower confidence.}
  \label{fig:monolayout_vis}
\end{figure*}

The MonoLayout-style learned baseline achieves the strongest overall localization performance among the evaluated BEV baselines on the aligned test subset. It reaches an ADE of 2.13\,m, FDE of 2.74\,m, ALE of 1.23\,m, ALgE of 1.49\,m, PCK@1m of 4.61\%, and PCK@2m of 28.16\%. In occupancy terms, it achieves 2.11 mIoU at threshold 0.25 and 1.06 mIoU at threshold 0.5. Although these occupancy numbers remain low, they show that dense monocular BEV reasoning is feasible under aerial supervision. Figure~\ref{fig:monolayout_vis} illustrates a representative qualitative result: the learned model recovers a smoother and more globally plausible BEV estimate, but the prediction remains diffuse relative to the drone-supervised target.

\begin{table}[!h]
  \caption{BEV baseline results on the aligned evaluation split. Coordinate-level metrics are reported in Table~\ref{tab:bev_coord_results}, and occupancy-level metrics are reported in Table~\ref{tab:bev_occ_results}.}
  \label{tab:bev_results}
  \centering

  \begin{subtable}{\columnwidth}
    \caption{Coordinate-level localization results.}
    \label{tab:bev_coord_results}
    \centering
    \begin{tabular}{lcccccc}
      \toprule
      Method & ADE $\downarrow$ & FDE $\downarrow$ & ALE $\downarrow$ & ALgE $\downarrow$ & PCK@1m $\uparrow$ & PCK@2m $\uparrow$ \\
      \midrule
      IPM baseline & 5.43 & -- & -- & -- & 0.00 & 6.70 \\
      BBox BEV regressor & 4.45 & -- & 1.68 & 3.77 & 3.50 & 13.00 \\
      MonoLayout-style baseline & 2.13 & 2.74 & 1.23 & 1.49 & 4.61 & 28.16 \\
      \bottomrule
    \end{tabular}
  \end{subtable}

  \vspace{0.6em}

  \begin{subtable}{\columnwidth}
    \caption{Occupancy-level BEV results.}
    \label{tab:bev_occ_results}
    \centering
    \begin{tabular}{lcc}
      \toprule
      Method & mIoU@0.25 $\uparrow$ & mIoU@0.5 $\uparrow$ \\
      \midrule
      MonoLayout-style baseline & 2.11 & 1.06 \\
      \bottomrule
    \end{tabular}
  \end{subtable}
\end{table}

\section{Conclusion}
We presented a cross-view urban traffic dataset and benchmark built from synchronized ego-centric bicycle videos and aerial drone videos recorded at real intersections. The benchmark supports two linked tasks: cross-view identity matching and ego-to-BEV prediction, together with a unified evaluation protocol and a reproducible pipeline for preprocessing, annotation, and baseline evaluation. By explicitly linking local ego observations with global aerial context, the benchmark fills a gap between cooperative perception datasets, cross-view retrieval benchmarks, and monocular BEV evaluation resources. Our baseline results show that the benchmark is both feasible and challenging. Cross-view matching attains strong recall but remains limited by over-assignment, temporal instability, and viewpoint ambiguity, while ego-to-BEV prediction highlights the gap between simple geometric projection and learned scene understanding in a lightweight monocular setting. These results indicate that the benchmark captures nontrivial open problems rather than being saturated by current baselines.

We hope this dataset will support future work on cross-view perception, urban scene alignment, and ego-to-global traffic understanding by providing a common test bed for studying how local observations can be linked to global spatial context in real urban environments. Beyond benchmark evaluation, the dataset may also support future research on intersection monitoring, traffic interaction analysis, and participant-density estimation under lightweight sensing constraints.

\bibliographystyle{plain}
\bibliography{ref}

\medskip

\newpage
\appendix

\section{Additional Dataset Statistics}
\label{app:dataset_stats}

This appendix provides additional dataset statistics complementing the summary reported in the main paper. In particular, we report the semantic composition of the dataset and a per-scene breakdown of frames, tracks, and validated cross-view correspondences.

As discussed in the main paper, the number of street-view track IDs is substantially larger than the number of drone-view track IDs because ego-centric tracking is more fragmented under motion blur, occlusion, and short object lifetimes. These counts should therefore be interpreted as tracker-level benchmark statistics rather than as direct counts of unique physical traffic participants.

\begin{table}[!h]
  \caption{Per-class statistics of the proposed dataset. Cars dominate the benchmark, but the dataset also contains pedestrians, bicycles, buses, trucks, and motorcycles, enabling evaluation across diverse urban traffic participants.}
  \label{tab:app_per_class_stats}
  \centering
  \begin{tabular}{lccc}
    \toprule
    Class & Street tracks & Drone tracks & Matched cross-view tracks \\
    \midrule
    Car         & 2,500 & 1,440 & 1,320 \\
    Truck       & 180   & 95    & 85 \\
    Bus         & 120   & 70    & 65 \\
    Person      & 550   & 240   & 210 \\
    Bicycle     & 420   & 220   & 180 \\
    Motorcycle  & 77    & 38    & 31 \\
    \midrule
    Total       & 3,847 & 2,103 & 1,891 \\
    \bottomrule
  \end{tabular}
\end{table}

The per-class statistics \ref{tab:app_per_class_stats} confirm that the benchmark is dominated by cars, as expected in urban traffic, while still containing pedestrians, cyclists, buses, and trucks. The per-scene breakdown further shows that the benchmark is not homogeneous across intersections: scenes differ in traversal length, participant density, and the proportion of track IDs that can be validated across views.

\section{Annotation Process and Anonymization}
\label{app:annotation}

A central contribution of the benchmark is the construction of identity-level cross-view correspondences between ego-centric street-view tracks and aerial drone-view tracks. Because these correspondences cannot be derived reliably from automatic matching alone, we construct them through a human-in-the-loop annotation workflow.

\subsection{Annotation Workflow}

\begin{figure*}[!h]
  \centering
  \includegraphics[width=\textwidth]{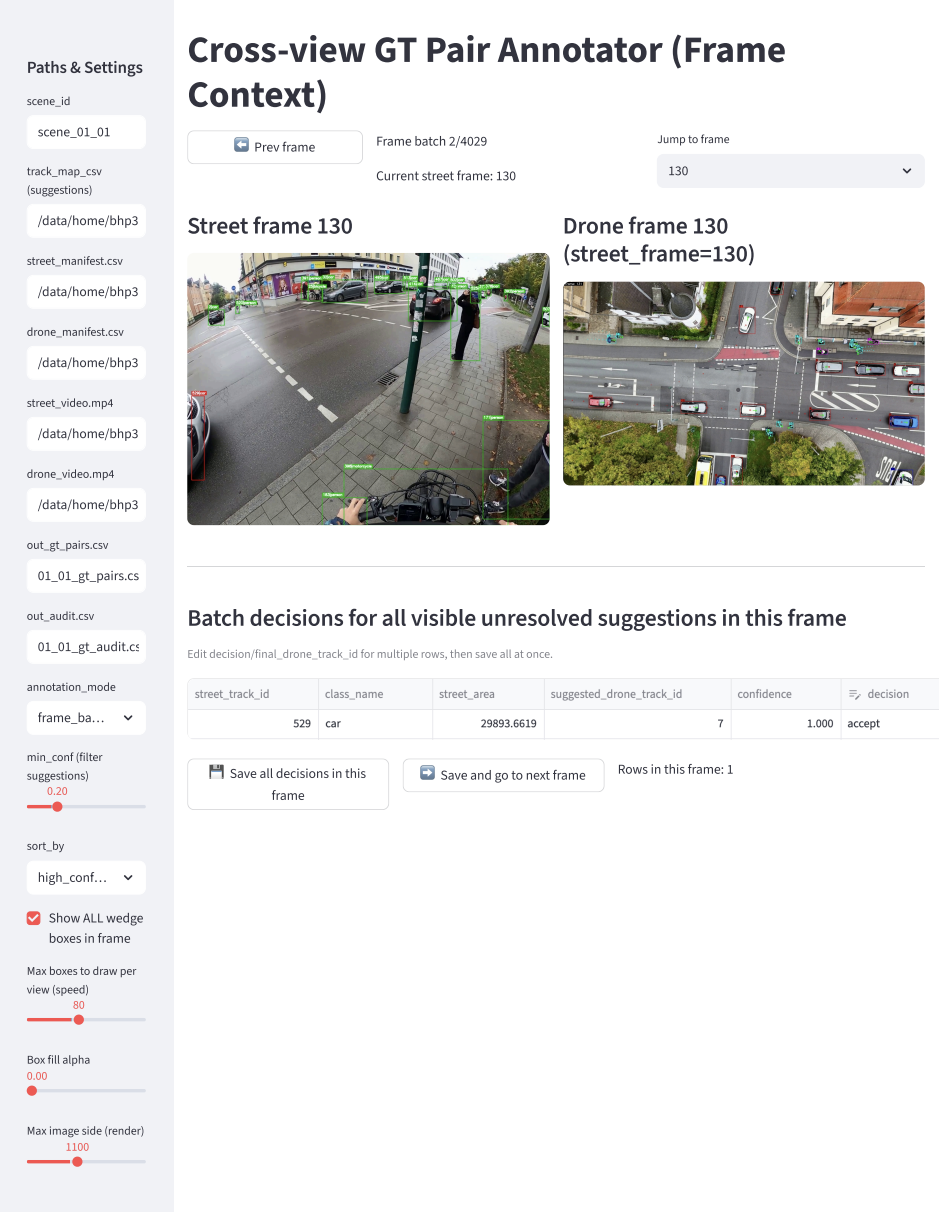}
  \caption{Web-based interface for ground-truth cross-view annotation. The UI shows synchronized street-view and drone-view imagery with tracked objects and candidate correspondences, allowing annotators to verify or reject matches and save the resulting pairs to \texttt{gt\_pairs.csv}.}
  \label{fig:app_annotation_ui}
\end{figure*}

The annotation process operates on synchronized street-view and drone-view videos together with per-view tracked detections. To make annotation scalable, we use a web-based interface (see Figure~\ref{fig:app_annotation_ui}) that supports both single-track verification and frame-batch verification. In frame-batch mode, annotators inspect all visible street-view tracks in a selected frame simultaneously and assign their corresponding drone-view track IDs or mark them as unmatched.

In practice, annotation proceeds as follows:
\begin{enumerate}
    \item A scene is selected from the scene manifest together with its synchronized street and drone views.
    \item The interface loads all currently visible street-view tracks and candidate drone-view tracks for a given aligned frame.
    \item For each visible street-view track, the annotator either assigns a drone-view track ID, marks the object as unmatched, or defers uncertain cases for later review.
    \item The verified assignments are stored in \texttt{gt\_pairs.csv}, while annotation history is optionally recorded in \texttt{gt\_audit.csv}.
\end{enumerate}

We found frame-batch annotation substantially faster than one-track-at-a-time verification because it allows the annotator to reason jointly about all visible objects in the scene, reducing repeated context switching and improving consistency.

\subsection{Ground-Truth Representation}

Ground-truth correspondences are stored in scene-specific \texttt{gt\_pairs.csv} files. Each row contains:
\begin{itemize}
    \item the scene identifier,
    \item the street-view track ID,
    \item the corresponding drone-view track ID,
    \item the semantic class label.
\end{itemize}

This representation makes the benchmark explicit at the track level while remaining compatible with frame-level evaluation, since track-level correspondences can be projected back onto synchronized frame pairs during evaluation.

\subsection{Why Human Verification Is Necessary}

Cross-view supervision in this setting is intrinsically difficult. Street-view tracks are often fragmented because of ego motion, partial visibility, occlusion, blur, and short track duration. By contrast, drone-view tracks are generally more stable because the aerial camera provides a wider and less occluded view of the full intersection. As a result, automatic matching alone is insufficient to produce reliable benchmark-quality supervision, especially for small or short-lived traffic participants.

The annotation process therefore serves two purposes: it makes the benchmark scientifically usable by providing verified correspondences, and it preserves a realistic level of difficulty by operating on tracker-generated identities rather than oracle trajectories.

\subsection{Anonymization}

To protect privacy and support public release, the dataset is anonymized before annotation and evaluation. In particular, ego-centric street-view videos are processed to obscure personally identifying visual information and applying gaussian blur for anonymization, including:
\begin{itemize}
    \item human faces,
    \item vehicle license plates.
\end{itemize}

Anonymization is applied before benchmark artifacts are released and before visual examples are selected for publication. This ensures that the released data preserve scene geometry, traffic context, and benchmark utility while reducing the risk of exposing personally identifying information.

\section{Additional Matching Results}
\label{app:matching_results}

This appendix provides additional diagnostic results for the cross-view matching benchmark beyond the aggregate tables reported in the main paper. We include macro and micro aggregation, scene-level variation, and class-level statistics.

\subsection{Macro and Micro Aggregation}

Table~\ref{tab:app_matching_macro_micro} reports macro and micro aggregation of the matching metrics. Macro averaging treats scenes equally, while micro averaging aggregates over all evaluated tracks and frames. The gap between these two summaries highlights scene imbalance and varying difficulty across intersections.

\begin{table}[!h]
  \caption{Macro and micro aggregation of cross-view matching results. Macro averages treat scenes equally, while micro averages aggregate over all evaluated tracks and frames.}
  \label{tab:app_matching_macro_micro}
  \centering
  \begin{tabular}{lccccccc}
    \toprule
    Aggregation & Track P & Track R & Track F1 & IDP & IDR & IDF1 & FrameAcc \\
    \midrule
    Macro & 33.7 & 100.0 & 50.2 & 64.9 & 29.5 & 40.5 & 27.0 \\
    Micro & 36.0 & 100.0 & 52.9 & 75.2 & 34.0 & 46.8 & 30.6 \\
    \bottomrule
  \end{tabular}
\end{table}

\subsection{Per-Scene Matching Results}

Table~\ref{tab:app_matching_per_scene_main}, ~\ref{tab:app_matching_per_scene_diag} reports scene-level matching performance for two scenes. The results show strong variation across intersections, which is consistent with differences in scene geometry, traffic density, and visibility conditions.

\begin{table*}[!h]
  \caption{Per-scene cross-view matching results. Scene-level variation is substantial, supporting the use of scene-level splits and scene-wise reporting for benchmark analysis.}
  \label{tab:app_matching_per_scene_main}
  \centering
  \small
  \begin{tabular}{lccccccc}
    \toprule
    Scene & Track P & Track R & Track F1 & IDP & IDR & IDF1 & FrameAcc \\
    \midrule
    scene\_01\_01 & 39.3 & 100.0 & 56.5 & 89.2 & 43.6 & 58.5 & 41.4 \\
    scene\_03\_01 & 28.1 & 100.0 & 43.9 & 40.6 & 15.5 & 22.5 & 12.7 \\
    \bottomrule
  \end{tabular}
\end{table*}

\begin{table*}[!h]
  \caption{Per-scene diagnostic breakdown of cross-view matching performance.}
  \label{tab:app_matching_per_scene_diag}
  \centering
  \small
  \begin{tabular}{lcccc}
    \toprule
    Scene & Near F1 & Far F1 & Stability & Consistency \\
    \midrule
    scene\_01\_01 & 51.2 & 77.9 & 40.0 & 94.5 \\
    scene\_03\_01 & 24.3 & 9.0 & 23.8 & 97.3 \\
    \bottomrule
  \end{tabular}
\end{table*}

\subsection{Per-Class Matching Results}

Table~\ref{tab:app_matching_per_class} reports per-class track-level matching performance. Larger rigid objects such as cars, buses, and trucks are generally easier to match, while pedestrians and cyclists remain more difficult because of their smaller size and more frequent occlusion in the street view.

\begin{table}[!h]
  \caption{Per-class track-level cross-view matching results. Larger rigid objects are generally easier to match, while pedestrians and cyclists remain more difficult.}
  \label{tab:app_matching_per_class}
  \centering
  \begin{tabular}{lcccccc}
    \toprule
    Class & P & R & F1 & TP & FP & FN \\
    \midrule
    Car         & 46.3 & 100.0 & 63.2 & 37 & 43 & 0 \\
    Truck       & 100.0 & 100.0 & 100.0 & 9 & 0 & 0 \\
    Bus         & 81.8 & 100.0 & 90.0 & 9 & 2 & 0 \\
    Person      & 57.1 & 100.0 & 72.7 & 12 & 9 & 0 \\
    Bicycle     & 55.6 & 100.0 & 71.4 & 10 & 8 & 0 \\
    Motorcycle  & 0.0 & 0.0 & 0.0 & 0 & 0 & 0 \\
    \bottomrule
  \end{tabular}
\end{table}

\subsection{Qualitative Matching Example}

\begin{figure*}[!h]
  \centering
  \includegraphics[width=\textwidth]{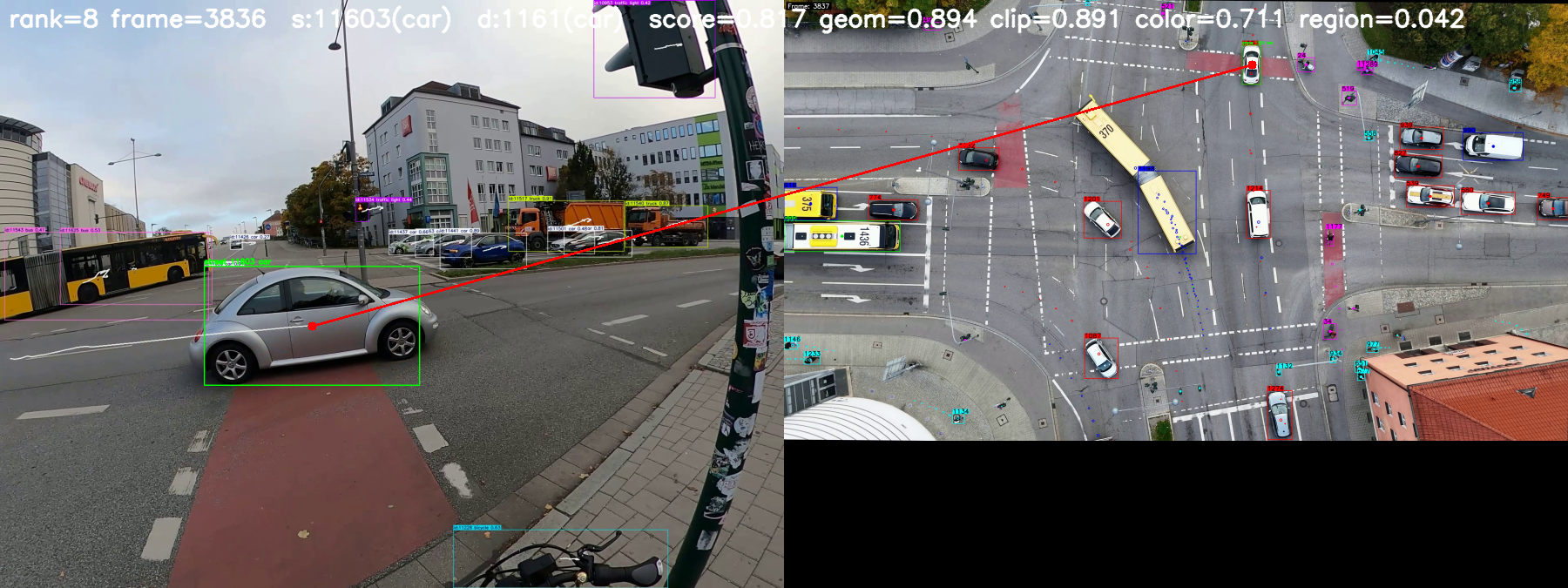}
  \caption{Qualitative example of cross-view identity matching. The left panel shows the ego-centric street view with the selected object highlighted, while the right panel shows the synchronized drone view with the matched aerial candidate. The visualization also reports the combined matching score together with component cues such as geometric consistency, CLIP similarity, color consistency, and regional compatibility. This example illustrates the large viewpoint gap addressed by the benchmark and the role of multi-cue matching in recovering cross-view correspondences.}
  \label{fig:app_matching_example}
\end{figure*}

Figure~\ref{fig:app_matching_example} shows a representative qualitative example of the cross-view matching baseline. The figure illustrates a matched street-view object and its associated drone-view candidate together with the component scores used by the matcher, including geometric consistency and CLIP similarity. Such examples help interpret why a correspondence is accepted and how appearance and geometry interact under strong viewpoint change.

\subsection{Formal Metric Definitions}

For completeness, we summarize the formal definitions of the matching metrics used in the benchmark.

\paragraph{Track-level evaluation.}
Let $\mathcal{G}$ denote the set of ground-truth cross-view correspondences and $\mathcal{P}$ the set of predicted correspondences. We define
\[
TP = |\mathcal{P} \cap \mathcal{G}|, \qquad
FP = |\mathcal{P} \setminus \mathcal{G}|, \qquad
FN = |\mathcal{G} \setminus \mathcal{P}|.
\]
Track-level precision, recall, and F1 are then
\[
P_{\mathrm{track}} = \frac{TP}{TP + FP}, \qquad
R_{\mathrm{track}} = \frac{TP}{TP + FN}, \qquad
F1_{\mathrm{track}} = \frac{2 P_{\mathrm{track}} R_{\mathrm{track}}}{P_{\mathrm{track}} + R_{\mathrm{track}}}.
\]

\paragraph{Frame-level evaluation.}
For each synchronized frame in which a ground-truth street track is visible, a method may assign a drone-view track ID. From the resulting frame-level assignments, we compute ID precision (IDP), ID recall (IDR), and IDF1:
\[
IDP = \frac{IDTP}{IDTP + IDFP}, \qquad
IDR = \frac{IDTP}{IDTP + IDFN}, \qquad
IDF1 = \frac{2 \cdot IDP \cdot IDR}{IDP + IDR}.
\]

\paragraph{Near--far split.}
To expose the geometric dependence of matching quality, we divide GT-visible frames into near and far subsets using the drone-side metric distance $d$:
\[
\text{near if } d \le \tau_{\mathrm{near}}, \qquad
\text{far if } d > \tau_{\mathrm{near}},
\]
where $\tau_{\mathrm{near}} = 30$\,m in our experiments.

\paragraph{Stability.}
For each street track with ground-truth $s$, stability is defined as the fraction of GT-visible frames in which the predicted drone ID matches the ground-truth drone ID:
\[
\mathrm{Stability}(s) = \frac{1}{T_s} \sum_{t=1}^{T_s} \mathbf{1}\{\hat{d}_{s,t} = d^*_{s}\},
\]
where $T_s$ is the number of GT-visible frames for street track $s$, $\hat{d}_{s,t}$ is the predicted drone ID at frame $t$, and $d^*_{s}$ is the ground-truth drone ID.

\paragraph{ID switches.}
We count the number of times the predicted drone ID changes along a street track:
\[
\mathrm{Switches}(s) = \sum_{t=2}^{T_s} \mathbf{1}\{\hat{d}_{s,t} \neq \hat{d}_{s,t-1}\}.
\]

\paragraph{Consistency.}
We also compute a track-level consistency score defined as
\[
\mathrm{Consistency}(s) = \frac{\max_d n_{s,d}}{\sum_d n_{s,d}},
\]
where $n_{s,d}$ is the number of frames in which the street track $s$ is assigned the drone ID $d$. A value of 1 indicates that the same predicted drone identity is used throughout all matched frames.

\subsection{Cross-View Matching Pipeline}

To reduce the search space and focus evaluation on the physically relevant region around the ego trajectory, the matching pipeline applies wedge filtering in BEV before candidate matching. The wedge is defined by the ego position and heading together with a configurable radius and opening angle. Only drone-side detections inside this region are considered as candidates for the current street-view frame.

\begin{figure*}[!h]
  \centering
  \includegraphics[width=\textwidth]{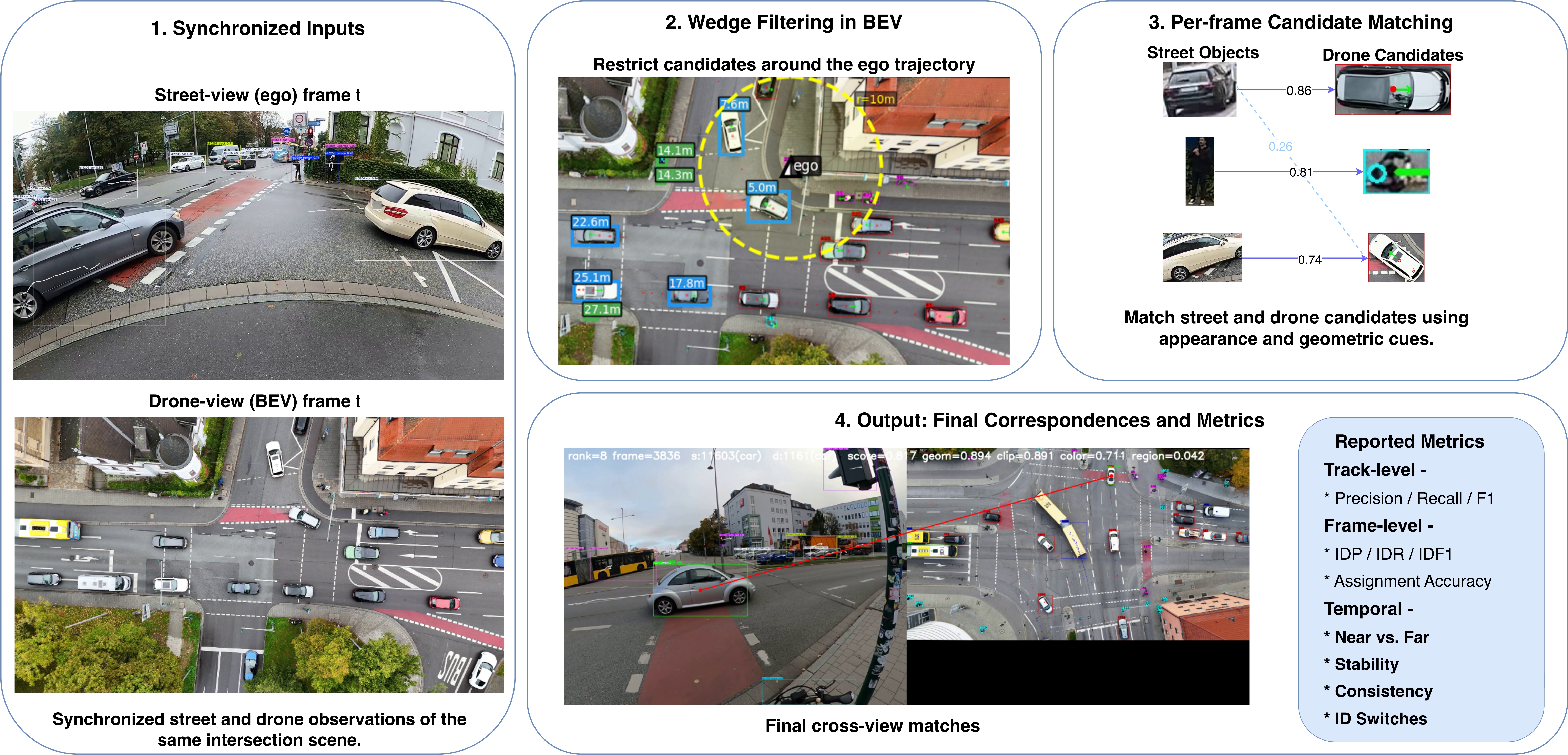}
  \caption{Cross-view matching sequence with wedge filtering. From left to right: synchronized street-view and drone-view inputs, wedge filtering in BEV to restrict candidate objects around the ego trajectory, frame-level matching using appearance and geometric cues, and final cross-view correspondences used for evaluation.}
  \label{fig:app_matching_pipeline}
\end{figure*}

Figure~\ref{fig:app_matching_pipeline} summarizes the full sequence from synchronized inputs to track-level correspondences and benchmark outputs.

\section{Implementation Details}
\label{app:implementation}

This appendix summarizes implementation details needed to reproduce the benchmark pipeline and baseline results.

\subsection{Scene Manifest and Preprocessing}

The benchmark is organized around a scene-level manifest. Each scene entry specifies:
\begin{itemize}
    \item street-view video path,
    \item drone-view video path,
    \item street-view detection/tracking CSV,
    \item drone-view detection/tracking CSV,
    \item ego track identifier in the drone view,
    \item ground-truth correspondence paths,
    \item processed output paths.
\end{itemize}

The scene manifest enables the full preprocessing pipeline to be rerun from scene-level inputs alone and supports batch evaluation over train/validation/test scene splits.

\subsection{Pipeline Structure}

The pipeline released follows the following sequence:
\begin{enumerate}
    \item scene manifest loading,
    \item wedge filtering around the ego trajectory,
    \item crop extraction for street and drone views,
    \item CLIP embedding computation,
    \item frame-level candidate matching,
    \item track-level temporal voting,
    \item human verification and evaluation.
\end{enumerate}

This organization supports both large-scale preprocessing and iterative annotation.

\subsection{Coordinate Alignment}

BEV evaluation requires reliable scene-level alignment between the street-view camera frame and the drone metric frame. We therefore compute per-scene coordinate alignment before evaluating BEV baselines. Scenes with unsuccessful alignment are marked accordingly and excluded from BEV training and evaluation unless explicitly enabled.

\subsection{BEV Baselines}

We evaluate three BEV baselines:
\begin{itemize}
    \item \textbf{IPM baseline}: projects street-view detections onto a ground plane using camera intrinsics and pitch assumptions, followed by scene-level coordinate alignment.
    \item \textbf{MonoLayout-style baseline}: predicts a dense BEV occupancy representation from the ego image using aerial supervision derived from aligned drone annotations.
    \item \textbf{BBox BEV regressor}: predicts metric BEV coordinates directly from object-level street-view bounding-box features and IPM priors.
\end{itemize}

\subsection{Training and Evaluation Settings}

All learned BEV baselines are trained and evaluated using scene-level splits. This avoids temporal leakage between train and test data and better reflects the intended generalization setting across intersections.

The current evaluation setup uses:
\begin{itemize}
    \item scene-level train/validation/test separation,
    \item aligned scenes only for BEV metrics,
    \item aggregate reporting across evaluation scenes,
    \item matching metrics at track, frame, geometric, and temporal levels,
    \item BEV metrics at both coordinate and occupancy levels where applicable.
\end{itemize}

\subsection{Code and Data Availability}
The dataset, annotation tools, preprocessing pipeline, and evaluation code will be made publicly available at \url{https://huggingface.co/datasets/prakharbh/CrossViewUrbanTrafficDataset} and the codebase is available at \url{https://github.com/oth-aifiud/Cross-View-Urban-Traffic-Dataset}.

\subsection{Repository Organization}

The repository separates:
\begin{itemize}
    \item \texttt{baselines/} for benchmark methods,
    \item \texttt{scripts/} for orchestration and preprocessing,
    \item \texttt{annotation/} for the web-based GT interface,
    \item \texttt{visualization/} for debugging and qualitative inspection.
\end{itemize}

Batch scripts are provided for:
\begin{itemize}
    \item matching evaluation,
    \item dataset statistics,
    \item coordinate alignment,
    \item IPM evaluation,
    \item MonoLayout-style training and evaluation,
    \item BBox regressor training and evaluation.
\end{itemize}

These components are intended to make the benchmark reproducible, extensible, and straightforward to evaluate at scene level.


\newpage

\end{document}